\documentclass[lettersize,journal]{IEEEtran}
\usepackage{amsmath,amsfonts}
\usepackage{algorithmic}
\usepackage{algorithm}
\usepackage{array}
\usepackage[caption=false,font=normalsize,labelfont=sf,textfont=sf]{subfig}
\usepackage{textcomp}
\usepackage{stfloats}
\usepackage{url}
\usepackage{verbatim}
\usepackage{graphicx}
\usepackage{cite}
\usepackage{soul,xcolor}
\usepackage{balance}

\begin{document}

\title{Integration of Agentic AI with 6G Networks for Mission-Critical Applications: Use-case and Challenges}

\author{Sunder Ali Khowaja$^{\ast}$,~\IEEEmembership{Senior Member,~IEEE,} Kapal Dev$^{\ast}$,~\IEEEmembership{Senior Member,~IEEE}, Muhammad Salman Pathan, Engin Zeydan,~\IEEEmembership{Senior Member,~IEEE}, Merouane Debbah,~\IEEEmembership{Fellow,~IEEE}
        % <-this % stops a space
\thanks{$^{\ast}$ Joint First Authors}
\thanks{Sunder Ali Khowaja is with School of Computing, Dublin City University and ADAPT Centre, Dublin, Ireland. Email: sunderali.khowaja@dcu.ie }% <-this % stops a space
\thanks{Kapal Dev is with Munster Technological University, Cork, Ireland. Email: kapal.dev@ieee.org}
\thanks{Muhammad Salman Pathan is with School of Computing, Dublin City University, Dublin, Ireland. Email: Muhammad.Salman@dcu.ie}
\thanks{Engin Zeydan is with Centre Tecnològic de Telecomunicacions de Catalunya, Castelldefels, Spain, 08860. Email: engin.zeydan@ctt.cat}
\thanks{ Merouane Debbah is with Khalifa University, Abu Dhabi, UAE. Email: merouane.debbah@ku.ac.ae}}

\markboth{Journal of \LaTeX\ Class Files,~Vol.~14, No.~8, August~2021}%
{Shell \MakeLowercase{\textit{et al.}}: A Sample Article Using IEEEtran.cls for IEEE Journals}


\maketitle

\begin{abstract}
We are in a transformative era, and advances in Artificial Intelligence (AI), especially the foundational models, are constantly in the news. AI has been an integral part of many applications that rely on automation for service delivery, and one of them is mission-critical public safety applications. The problem with AI-oriented mission-critical applications is the "human-in-the-loop" system and the lack of adaptability to dynamic conditions while maintaining situational awareness. Agentic AI (AAI) has gained a lot of attention recently due to its ability to analyze textual data through a contextual lens while quickly adapting to conditions. In this context, this paper proposes an AAI framework for mission-critical applications. We propose a novel framework with a multi-layer architecture to realize the AAI. We also present a detailed implementation of AAI layer that bridges the gap between network infrastructure and mission-critical applications. Our preliminary analysis shows that the AAI reduces initial response time by 5.6 minutes on average, while alert generation time is reduced by 15.6 seconds on average and resource allocation is improved by up to 13.4\%. We also show that the AAI methods improve the number of concurrent operations by 40, which reduces the recovery time by up to 5.2 minutes. Finally, we highlight some of the issues and challenges that need to be considered when implementing AAI frameworks. 
\end{abstract}

\begin{IEEEkeywords}
Agentic AI, mission critical services, 6G networks, use cases.
\end{IEEEkeywords}

\section{Introduction}

\begin{figure}[!t]
\centering
\includegraphics[width=\linewidth]{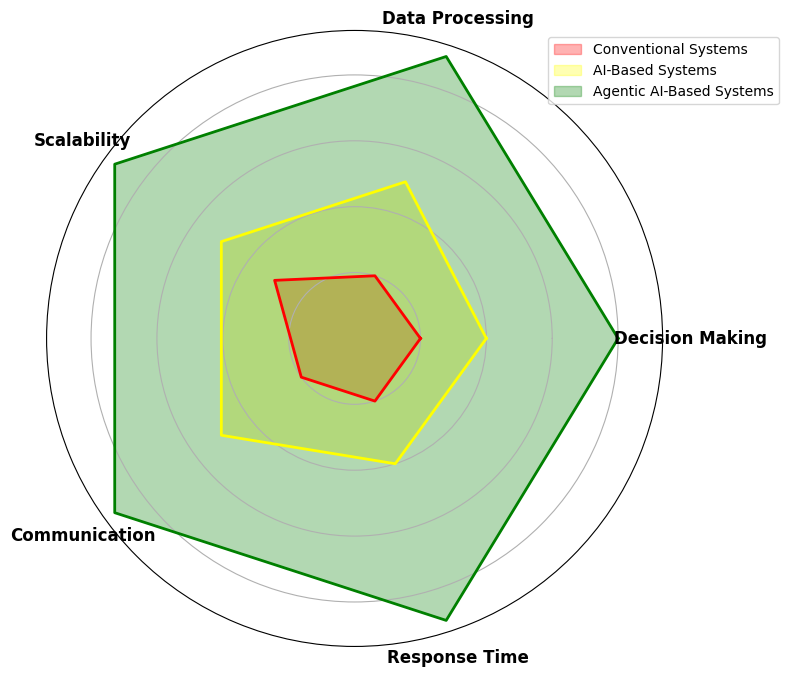}
\caption{Radar chart comparing conventional, AI-based and AAI-based mission-critical public safety applications in terms of response time, communication, scalability, data processing and decision making. }
\label{fig_1}
\end{figure}

Mission-critical public safety applications require collaboration between law enforcement agencies, advanced technological frameworks that enable real-time decision making, emergency response teams and adaptive intelligence. Such systems pose a challenge as they are associated with high risks. These include computational constraints, communication limitations in dynamic threat landscapes, incomplete information and time urgency \cite{6GMC, SmartMC}. The current generation of mission-critical public safety systems uses human-supervised decision-making processes with centralized command and control infrastructures and rules-based automation systems. While these systems have proven successful in controlled or well-defined operational scenarios, they reach their limits when the environment is dynamic, complex and unpredictable \cite{IMC5G, SmartMC}. The complex and dynamic situations are accompanied by challenges resulting from delays in data processing pipelines, lack of real-time situational awareness, rapid deployment and optimization of resource allocation. In addition, most traditional mission-critical public safety applications face a single point of failure, creating bottlenecks in the seamless processing of concurrent scenarios. It should also be noted that the rule-based systems that are mostly used for mission-critical applications are not adaptive and therefore do not respond well to the situation that is not pre-programmed \cite{SmartMC, 6GMC, IMC5G}.\\
To overcome the problems of pre-programming rule-based systems, artificial intelligence (AI)-based mission-critical systems have been proposed to improve mission-critical public safety applications \cite{AIMC}. AI systems leverage natural language processing, computer vision and machine learning techniques to develop decision support systems that can perform real-time analysis. AI models help human operators to process huge amounts of sensor data, automate incident detection and recommend informed decisions. For instance, AI-based systems can help predict disaster areas through predictive analytics, and computer vision systems help analyze surveillance imagery for real-time threat detection. The AI-based mission-critical applications have greatly improved in terms of performance compared to the rule-based systems, but still act as a tool for decision-making and recommendations rather than autonomous entities. Such systems still require human oversight and rely on predefined models within a centralized ecosystem \cite{SmartMC, 6GMC, IMC5G}. They also work reactively, i.e. they only respond when prompted and follow strict, predefined patterns. Furthermore, the problems of limited adaptability, processing delay constraints, dealing with dynamic situations and collaboration between multiple AI systems in complex scenarios remain the same.\\
With the rapid emergence of large language models (LLMs), a fundamental paradigm shift can be observed in the use of AI methods for multimodal data. This paradigm shift is observed in the form of Agentic AI (AAI) \cite{AgenticAI}. AAI refers to AI agents that act autonomously based on contextual information to achieve specific goals. In contrast to conventional AI systems, we assume that AAI-enabled systems can improve the performance of mission-critical operations by deploying autonomous, context-aware and self-improving AI agents. AAI will enable the system to actively adapt, reason and perceive dynamic conditions in real time. In addition, AAI can be used with next-generation communication systems such as edge intelligence, software-defined networking, and 5G/6G networks to perform distributed decision making \cite{6GAI}. The fusion of novel technologies with AAI will help reduce reliance on centralized processing, thereby reducing latency while improving operational resilience. \\
Based on the above assumptions, we can give an example of a large-scale emergency scenario such as urban security threat or a natural disaster, where information from different data streams and human supervisors are needed to coordinate responses and interpret data. For this case, let's consider a severe weather event. The traditional mission-critical system would require AI systems to collect and analyze the data from various sources. However, the human operator would still need to coordinate with the emergency response units and manage resource allocation. Such a scenario leads to bottlenecks, miscommunication and inefficiencies in the mobilization of resources. The use of AI can automate the data analysis and recommendation part, but a human supervisor is still required to make the final decision and mobilize the resources. With AAI-driven systems, agents can autonomously analyze data from multimodal sensors, predict the progression of hazards, dynamically assign responders and coordinate with cross-agency facilities in real time. Decentralizing AI systems to autonomous agents helps mission-critical systems operate at the edge, which can improve responsiveness, scalability and agility while ensuring effective and faster intervention. We illustrate a hypothetical comparison between conventional AI-based and AAI-based mission-critical public safety applications in Figure \ref{fig_1} in terms of key characteristics such as response time, communication, scalability, data processing and decision making. Conventional systems are only adaptable to a limited extent, react slowly and make their decisions manually. The AI-based system improves the efficiency of responsiveness and decision-making, but still requires human oversight. However, the AAI is autonomous, adaptable and has the ability to make decisions in real time and with high scalability. \\

\section{Public Safety Networks}

Mission-critical public safety systems are specialized applications designed to ensure that essential emergency services remain operational to protect communities \cite{Ulema}. Traditionally, these systems were limited to voice and short message communication over narrowband networks. However, the increasing complexity and unpredictability of modern public safety scenarios have driven the adoption of advanced technologies such as AI, edge computing and high-speed networks such as 5G/6G, transforming these systems into decentralized, autonomous platforms capable of making real-time decisions and improving situational awareness. As a result, modern systems are becoming more adaptable and resilient and are able to process large amounts of data in real time, ultimately improving their ability to protect lives and assets in critical situations.

There are several papers in which researchers have explored the integration of 6G and AI to improve mission-critical applications. Authors in \cite{6GMC} investigated the role of Mission-Critical Internet of Things (MC-IoT) in 6G networks, focusing on its importance for public safety and emergency response.  The paper in \cite{IMC5G} presents a proactive solution for overload detection and resource scaling of Mission Critical Services (MCS) in Beyond 5G (B5G) networks based on an intelligent loop between the MCS server, Network Data Analytics Function (NWDAF) and orchestrator using AI-based real-time inference for predictive scaling.  Kalør et al \cite{Wireless6G}  discusses the evolution of massive machine type communications (mMTCs) and ultrareliable low-latency communications (URLLCs) under the umbrella of 6G to support a variety of safety-critical control applications with different types of timing requirements, as evidenced by the emergence of metrics related to information freshness and information value.  Gupta et al. \cite{6Gedge} present a 6G-driven Edge Intelligence (6G-EI) framework and emphasize its potential for URLLC in mission-critical applications such as autonomous vehicles, border surveillance, and tele-surgery.  The authors demonstrate the effectiveness of 6G-connected EI using a UAV-based pandemic response case study and compare 4G, 5G and 6G in terms of latency, security and mobility. Chataut et al \cite{6GAI} discuss mission-critical applications of 6G, including public safety, autonomous systems and industrial IoT and explore key technologies such as terahertz communications, ultra-massive MIMO, reconfigurable intelligent surfaces (RIS) and quantum communications that will drive the development of 6G. Wang et al. \cite{XAI6G} investigate the integration of XAI into 6G networks, focusing on improving trust, transparency and interpretability in AI-driven decision-making processes and can improve trust and accountability in AI-powered mission-critical systems, particularly in healthcare, smart cities and emergency response.    Spantideas et al. \cite{SmartMC} present an AI-powered management framework for MCS that leverages ML-driven overload detection, network slicing and edge computing for improved emergency response.

In this paper, we propose a theoretical, multi-layered architecture and relevant technologies for the deployment of AAI in mission-critical public safety applications. We provide a comprehensive analysis of the framework that utilises AAI to improve automated decision making and situational awareness for operational efficiency and emergency response, respectively. We also address critical considerations of ethical implications, operational trust, and system security related to the use of AAI in public safety. Finally, we list the lessons learned from the study and conclude the study with some future extensions of this work.

\section{Proposed Architecture}

\begin{figure}[!t]
\centering
\includegraphics[width=\linewidth]{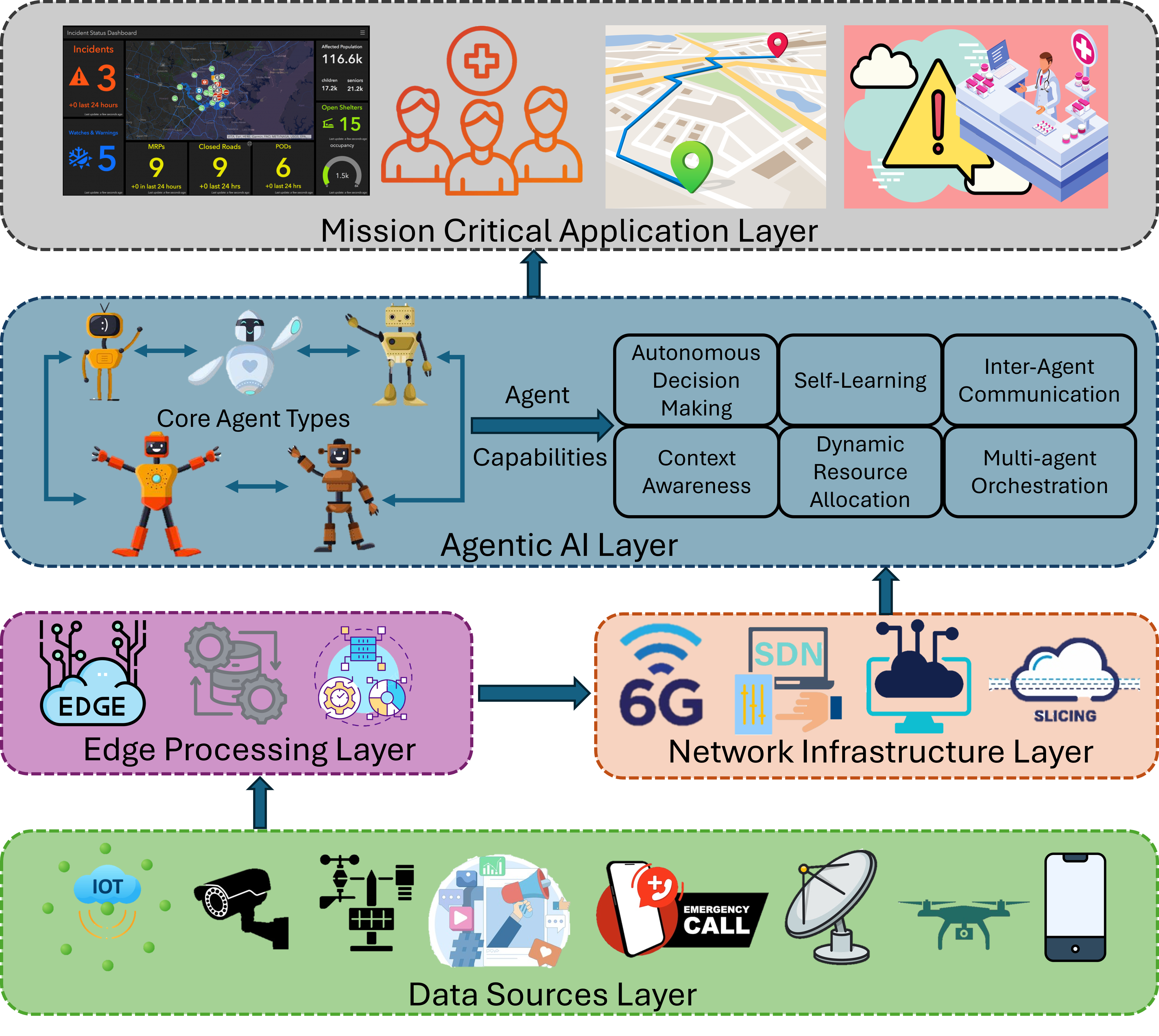}
\caption{Proposed Layered Architecture for AAI-based mission critical applications for public safety}
\label{fig_2}
\end{figure}

The proposed architecture for AAI-based mission-critical public safety applications is shown in Figure \ref{fig_2}. We consider a disaster management use case to illustrate the layers and architectural components that enable efficient resource allocation, adaptive response coordination, and real-time decision making in mission-critical scenarios. The proposed system integrates sensors and IoT devices, distributed AI agents, advanced communication infrastructure, edge computing and cloud-based orchestration to create an autonomous and resilient framework for disaster response. The architecture is modular and consists of several independent layers, each of which contributes to the efficiency and effectiveness of the overall system. In the following subsections, we explain each layer in detail.

\subsection{Data Sources Layer}

At the foundational level, mission-critical applications require data to be collected from a variety of devices. Therefore, the data source layer is designed to capture data from a variety of sensors and IoT-based devices, including but not limited to social media analytics, unmanned aerial vehicles, surveillance cameras, flood detectors, temperature monitors, seismic sensors and more. The collection of data from heterogeneous sources enables the system to provide situational awareness while ensuring a comprehensive understanding of the scenario. The sensor networks used for an existing application, such as smart city infrastructure and remote sensing satellites, can also be leveraged to serve as the basis for this layer. The integration of mobile crowdsourcing platforms with the existing sensor data would improve data granularity while enabling rapid updates and dynamic response mechanisms.

\subsection{Edge Processing Layer}

The edge processing layer is responsible for decentralized data processing, analysis, fusion and decision making in real time. We can also deploy AI agents at this layer that process the data locally on edge computing nodes such as multi-access edge computing nodes, IoT gateways and unmanned aerial vehicles to reduce latency and improve response time. We can also utilise distributed learning paradigms such as federated learning, split learning or split- federated learning to ensure that the edge computing nodes have on-device AI capabilities for adaptive learning without over-reliance on the cloud computing platforms \cite{DBFL}. Such computing capacities and functionalities are already available in the form of NVIDIA Jetson platforms and Google Coral AI accelerators. These are just a few examples of many available devices that have similar capabilities to perform the required computation. The distributed AI agents can filter out the noise from the sensor data, detect anomalies and even initiate initial emergency measures, such as rerouting communication paths or broadcasting alerts in advance of infrastructure failures.

\subsection{Network Infrastructure Layer}

Once the edge nodes perform local computation for data fusion and analysis, the network infrastructure layer would facilitate data transmission with low latency and in a robust manner through 5G, 6G and software defined networks (SDN) \cite{Wireless6G}. The network infrastructure layer will also be responsible for connecting edge device computation, decision making and analysis to the AAI layer in a robust manner. The specific applications for mission-critical services require URLLC, which can be achieved through adaptive routing and network slicing with respect to the above-mentioned communication medium. Next generation communication systems, including Public Safety LTE (PS-LTE) and FirstNet \cite{PublicSafetyLTE, FirstNet}, implicitly provide emergency communication capabilities. However, if such next-generation networks are equipped with AI-driven adaptive network functions, resilience for mission-critical services can be further improved. Network function virtualization can ensure dynamic reconfiguration of connectivity in the event of a network infrastructure failure, enabling seamless coordination between central command centers and agents in the field. 

\begin{figure*}[!t]
\centering
\includegraphics[width=\linewidth]{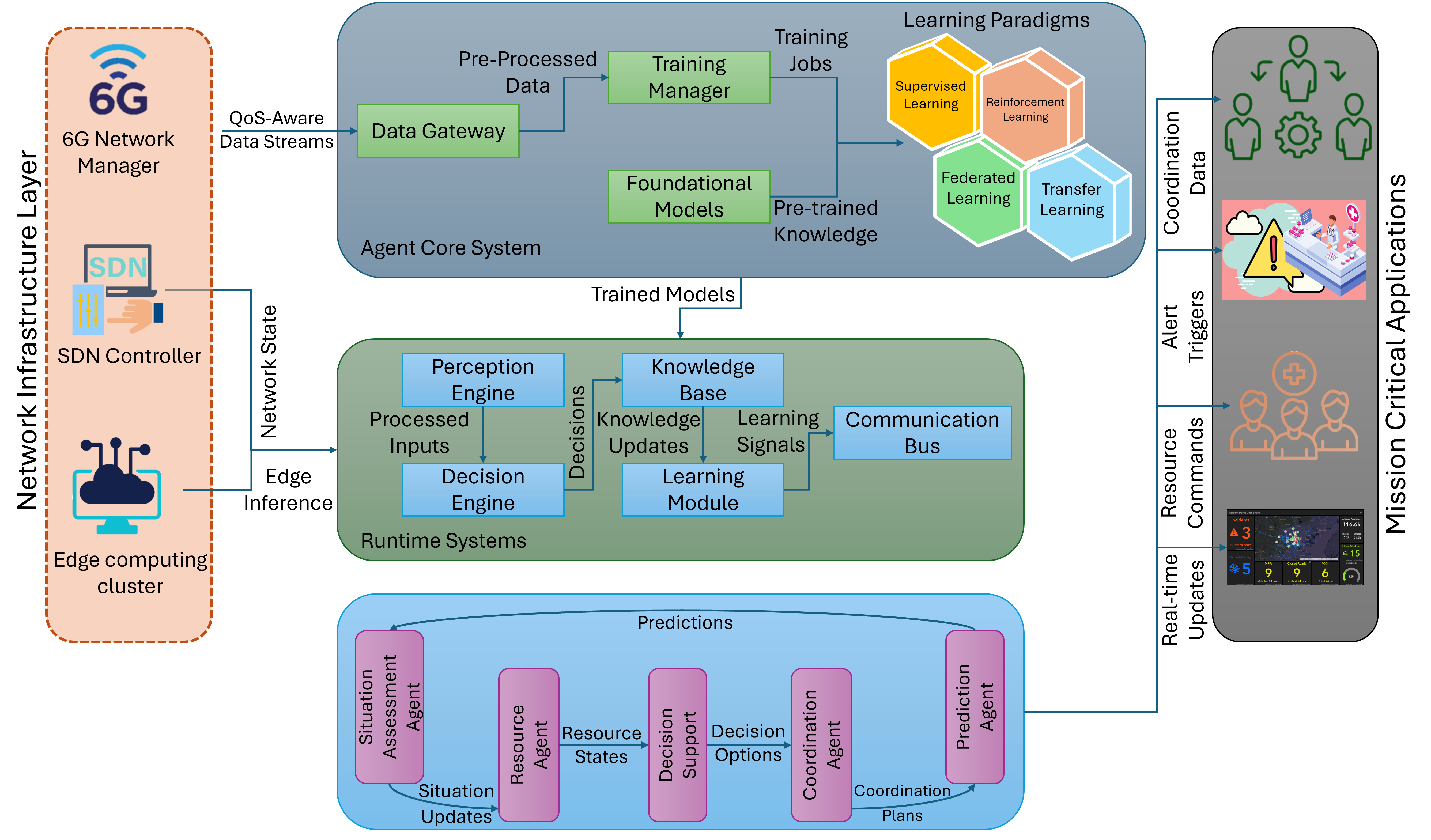}
\caption{The in-depth architectural details for AAI layer.}
\label{fig_3}
\end{figure*}

\subsection{Agentic AI Layer}

The backbone of the proposed architecture for mission-critical applications is the AAI layer, which enables collaborative operations, continuous learning and autonomous decision making between distributed AI entities. In contrast to traditional AI-based techniques, AAI provides adaptive intelligence that enables the proposed system to dynamically adapt to the situation at hand. The detailed architecture of the AAI layer is shown in Figure \ref{fig_3}.\\
The figure shows that the network infrastructure layer is the input for the AAI layer, which ensures real-time connectivity and low-latency processing of the data or decisions received from the edge processing layer. The network infrastructure also helps AAI layer to support dynamic bandwidth allocation and slicing capabilities to ensure seamless data transmission across mission critical applications. The Agentic core system receives the data from the Data Gateway, which is then sent to the training manager and the foundational models for further processing. The Training Manager and the Foundational Manager can use different learning paradigms such as transfer learning, federated learning, reinforcement learning or supervised learning to improve the decision-making process. In addition, the pre-trained models are constantly updated in the context of the application to improve the situational awareness of the mission-critical application. Some of the foundational models available include the model hub from HuggingFace Transformers, OLLAMA and other similar platforms. The runtime systems use a microservices architecture with service mesh capabilities to run a perception engine that forwards the information from the agent core system to the decision engine. The knowledge base leverages knowledge from past mission critical deployment scenarios to help the learning module refine decision-making strategies. For example, the knowledge base can be represented by a distributed graph to store operational states and domain knowledge. In addition, operational feedback can also be stored in a knowledge base to enable continuous adaptation of the model. A communication bus is provided for coordination between the AI agents and the emergency response teams, but this can be modified according to the specific requirements of the operators and system designers. For example, Apache Pulsar can be used for communication, ensuring reliable delivery of messages and providing geo-replication capability and multi-tenancy support.\\
Finally, the multi-agent system utilizes various AI agents, which include but are not limited to, situation assessment agents, resource agents, decision support agents, coordination agents and prediction agents. Each specialized agent can undergo a multi-phase training process and can operate on multimodal or unimodal data. The agents work collectively to process situation updates, optimize resource allocation, generate decision options, coordinate emergency response, and predict the course of the disaster. The output of the multi-agent system directly influences the reaction mechanisms at field level. The decisions of the multi-agent system are then forwarded to the mission-critical application layer. 

\subsection{Mission Critical Application Layer}

The mission-critical application layer acts as an interface between the AAI layer and the disaster response efforts. The functionalities performed at this layer include but not limited to decision support systems, resource allocation, risk assessment and emergency coordination, which have a direct impact on public safety measures. With the integration of AAI layer,  the application layer can become more efficient, adaptable and responsive. The use of AAI also enables seamless collaboration between autonomous AI agents and emergency responders as they receive the recommendations from the AI agents directly to their decision support interface. The emergency responders can then validate, override or modify the proposed actions, which in turn helps the AI agents to learn adaptively and learn and respond better in the next update. 

\section{Experimental Setup and Analysis}

In this section, we first present the dataset used to evaluate the proposed approach, then the experimental setup and finally the analysis.

\subsection{Dataset Description and Preprocessing}

For the realization of the proposed AAI architecture for mission-critical applications, we used the Federal Emergency Management Agency (FEMA) disaster declarations summary dataset (1953-2023)\footnote{https://www.fema.gov/openfema-data-page/disaster-declarations-summaries-v2}, combined with National Oceanic and Atmospheric Administration (NOAA) Storm Events database\footnote{https://www.ncdc.noaa.gov/stormevents/details.jsp}. The combined dataset includes approximately 58.4K disaster events categorized into wildfires, severe storms, hurricanes and floods. We also include additional data from the CrisisLexT26 dataset \cite{Crisis}, which contains around 250K tweets on 26 crisis events. To aggregate the data from different sources, we perform a temporal alignment with timestamps, a standardization of the coordinate system and feature engineering. We used the MICE algorithm to process the missing data and obtain temporal dependencies.

\subsection{Experimental Setup}

For the experimental analysis, we develop a scaled-down version of the proposed AAI architecture that focuses on only three agents, namely agents for situation assessment, resource management, and prediction. We used the edge nodes using 3 AWS t3.xlarge instances with NVIDIA GPUs, the central processing was performed with 1 AWS p3.2xlarge instance with NVIDIA GPU, and the network was simulated with the NS3 network simulator according to the 5G network conditions. We implemented the proposed method using the PyTorch 2.0 framework. We used RLlib 2.4.0 for the agent framework, the knowledge base was based on Neo4j and Apache Kafka was used as a message broker. We partitioned the data for the training period, the validation period and the test period into the years 1953-2010, 2011-2018 and 2019-2023, respectively. We compare our approach with the rule-based AI using ICS (Incident Command System) of FEMA, LSTM, Transformer network \cite{Transformer}, and the proposed AAI method, respectively. We present the evaluation with metrics such as initial response time, decision-making latency, alert generation time. We also perform a situation assessment, which is system classification against ground truth post-incident reports, an optimal resource distribution index, which is the ratio between correctly allocated resources and total resources, and response time efficiency, that is the ratio between baseline response time and actual response time. To show the efficiency, we also provide the average bandwidth consumption during active incidents, memory usage and CPU utilization evaluated with AWS CloudWatch.

\subsection{Implementation Details}

%For the implementation of this work, we implemented three agents, i.e. agents for situation assessment, resource management and prediction. 

For the situation assessment agent, we used the base BERT model for text processing and ResNet50 for analyzing images. We trained the data on 70\%, while the remaining data was used equally for validation and testing. We used the AdamW optimizer with default parameters. The batch size was set to 32 and the learning rate to $2e^{-5}$. For the resource management agent, we used PPO with a custom reward function to optimize resource usage and response time. We used a 24-dimensional vector that includes deployment status and resource availability. We used discrete actions for resource allocation decisions as the action space and trained the network with 10,000 training episodes, opting for early stopping if reward convergence is achieved. For the prediction agent, we used a probabilistic neural network with uncertainty estimation whose 32-dimensional input features consist of disaster characteristics vector. The output of the prediction agent was multi-step prediction of disaster progression along with confidence intervals.

\begin{table*}[]
\centering
\caption{Comparative analysis between the proposed Agentic AI versus non-Agentic AI approaches across number of performance metrics}
\label{tab:my-table}
\begin{tabular}{|ccccc|}
\hline
\multicolumn{1}{|c|}{\textbf{Performance Metric}} &
  \multicolumn{1}{c|}{\textbf{Rule-based}} &
  \multicolumn{1}{c|}{\textbf{LSTM}} &
  \multicolumn{1}{c|}{\textbf{Transformer}} &
  \textbf{Agentic AI} \\ \hline
\multicolumn{1}{|c|}{\textbf{Response time}}         & \multicolumn{1}{c|}{8.8 m}    & \multicolumn{1}{c|}{8.2 m}    & \multicolumn{1}{c|}{8.6 m}    & 3.2 m    \\ \hline
\multicolumn{1}{|c|}{\textbf{Decision Latency}}      & \multicolumn{1}{c|}{28.6 s}   & \multicolumn{1}{c|}{18.4 s}   & \multicolumn{1}{c|}{16.5 s}   & 8.9 s    \\ \hline
\multicolumn{1}{|c|}{\textbf{Alert Generation}}      & \multicolumn{1}{c|}{22.4 s}   & \multicolumn{1}{c|}{13.3 s}   & \multicolumn{1}{c|}{12.1 s}   & 6.8 s    \\ \hline
\multicolumn{1}{|c|}{\textbf{Situation Assessment}}  & \multicolumn{1}{c|}{85.2 \%}  & \multicolumn{1}{c|}{86.4 \%}  & \multicolumn{1}{c|}{86.8 \%}  & 94.0 \%  \\ \hline
\multicolumn{1}{|c|}{\textbf{Resource Allocation}}   & \multicolumn{1}{c|}{75.8 \%}  & \multicolumn{1}{c|}{78.9 \%}  & \multicolumn{1}{c|}{80.1 \%}  & 89.2 \%  \\ \hline
\multicolumn{1}{|c|}{\textbf{Accuracy}}              & \multicolumn{1}{c|}{79.4 \%}  & \multicolumn{1}{c|}{82.3 \%}  & \multicolumn{1}{c|}{84.0 \%}  & 88.1 \%  \\ \hline
\multicolumn{1}{|c|}{\textbf{False Alarm rate}}      & \multicolumn{1}{c|}{8.6 \%}   & \multicolumn{1}{c|}{7.8 \%}   & \multicolumn{1}{c|}{7.5 \%}   & 4.2 \%   \\ \hline
\multicolumn{1}{|c|}{\textbf{CPU Utilization}}       & \multicolumn{1}{c|}{68.4 \%}  & \multicolumn{1}{c|}{72.3 \%}  & \multicolumn{1}{c|}{75.6 \%}  & 82.6 \%  \\ \hline
\multicolumn{1}{|c|}{\textbf{Memory Usage}}          & \multicolumn{1}{c|}{8.6 GB}   & \multicolumn{1}{c|}{9.8 GB}   & \multicolumn{1}{c|}{11.2 GB}  & 12.4 GB  \\ \hline
\multicolumn{1}{|c|}{\textbf{Network Bandwidth}}     & \multicolumn{1}{c|}{42.8 Mbs} & \multicolumn{1}{c|}{62.4 Mbs} & \multicolumn{1}{c|}{64.6 Mbs} & 68.2 Mbs \\ \hline
\multicolumn{1}{|c|}{\textbf{System Availability}}   & \multicolumn{1}{c|}{99.92 \%} & \multicolumn{1}{c|}{99.95 \%} & \multicolumn{1}{c|}{99.95 \%} & 99.98 \% \\ \hline
\multicolumn{1}{|c|}{\textbf{Recovery Time}}         & \multicolumn{1}{c|}{8.6 m}    & \multicolumn{1}{c|}{6.4 m}    & \multicolumn{1}{c|}{6.6 m}    & 3.4 m    \\ \hline
\multicolumn{1}{|c|}{\textbf{Concurrent Operations}} & \multicolumn{1}{c|}{45}       & \multicolumn{1}{c|}{67}       & \multicolumn{1}{c|}{64}       & 85       \\ \hline
\multicolumn{5}{|c|}{\textbf{Specific Downstream tasks}}                                                                                                        \\ \hline
\multicolumn{1}{|c|}{\textbf{Flood Prediction}}      & \multicolumn{1}{c|}{82.6 \%}  & \multicolumn{1}{c|}{85.7 \%}  & \multicolumn{1}{c|}{86.2 \%}  & 92.4 \%  \\ \hline
\multicolumn{1}{|c|}{\textbf{Hurricane Prediction}}  & \multicolumn{1}{c|}{78.4 \%}  & \multicolumn{1}{c|}{81.3 \%}  & \multicolumn{1}{c|}{81.6 \%}  & 87.9 \%  \\ \hline
\multicolumn{1}{|c|}{\textbf{Wildfire Prediction}}   & \multicolumn{1}{c|}{74.8 \%}  & \multicolumn{1}{c|}{72.3 \%}  & \multicolumn{1}{c|}{78.7 \%}  & 85.6 \%  \\ \hline
\end{tabular}
\label{tab:table_1}
\end{table*}

\begin{table}[]
\centering
\caption{Qualitative Comparison of System Characteristics between Agentic and non-Agentic AI approaches}
\label{tab:my-table}
\begin{tabular}{|c|c|c|c|c|}
\hline
\textbf{Characteristic} & \textbf{Rule-based} & \textbf{LSTM} & \textbf{Transformer} & \textbf{Agentic AI} \\ \hline
\textbf{Adaptability}  & Mod   & Mod   & Mod   & High \\ \hline
\textbf{Learning}      & Lim   & Lim   & Lim   & Adv  \\ \hline
\textbf{Scalability}   & Low   & Mod   & Mod   & High \\ \hline
\textbf{Edge handling} & Low   & Mod   & Mod   & High \\ \hline
\textbf{Coordination}  & SAuto & SAuto & SAuto & Auto \\ \hline
\textbf{Real-time}     & Low   & Mod   & Mod   & High \\ \hline
\textbf{Uncertainty}   & None  & Lim   & Lim   & High \\ \hline
\textbf{Multimodality} & Lim   & Mod   & Mod   & Adv  \\ \hline
\end{tabular}
\label{tab:table_2}
\end{table}

\subsection{Experimental analysis}

The aim of this work is to show a basic realization of the proposed AAI framework and to demonstrate its effectiveness in comparison to rule-based, LSTM-based and transformer-based networks. We report the results in Table \ref{tab:table_1}. It can be seen that the proposed AAI method reduces the response time by more than half of what is achieved with non-agentic methods. This improvement is due to parallel processing capabilities, the use of human-in-the-loop and edge computing deployment. Accuracy has also been improved by the AAI approach, as it has the ability to fuse multimodal data, learn and adapt in real time and improve pattern recognition. Although the AAI method performs well overall, it definitely requires more memory and bandwidth, which is due to multiple AI models running simultaneously and continuously learning with model updates. These simulations were performed with a 5G scenario, so we assume that there would not be the same limitations in 6G networks. In addition, we perform a qualitative comparison of the system characteristics observed by the experimental analysis. The comparison is shown in Table \ref{tab:table_2}. Based on the analysis, it is assumed that the AAI-based method is better in the areas of adaptability, learning ability, handling edge cases, real-time decision making, handling uncertainty, and multimodal data processing, respectively. The terms Mod, Lim, SAuto, Auto and Adv in Table \ref{tab:table_2} correspond to the terms Moderate, Limited, Semi Automated, Automated and Advanced respectively.

\section{Open Research Issues and Challenges}

Although AAI has advantages over non-agentic methods, there are some open research questions and challenges that are highlight in this section.

\subsection{Openness of Foundational Model and Adaptability}

AAI methods rely heavily on the foundational models for mission-critical applications. However, most of the foundational models with the best performance are closed-source or restricted. Furthermore, state-of-the-art large language models and vision transformers are trained on proprietary datasets that are not transparent in terms of data provenance, failure modes, and biases. This not only makes it more difficult to fine-tune models for certain scenarios, but also limits the audit of the decision-making process. Furthermore, the problem of interoperability also arises when different AI agents use different LLMs. More interoperable open source foundational models that can  be available in the future can reduce this potential limitation. 

\subsection{Trust, Interpretable, and Accountability}

As the AAI systems introduce a high degree of automation so that the autonomous agents automatically assess situations and make recommendations, the human responders might have difficulty with this transition  to follow the instructions generated by the AI due to potential trust issues. Another problem can be interpretability, as emergency responders would need a clear justification for the generated decisions in order to act accordingly. Since most of the available LLMs are black boxes, it is difficult to understand the decision logic. It is likely that without transparency, responders may hesitate to follow AI-based instructions, delaying critical actions. It is therefore crucial to bridge the gap between trust and interpretation. Accountability will also be a pressing concern when it comes to AAI, as the systems that need to be designed should be optimal. In addition, the design should follow an accountability framework that includes AI audit trails, legal responsibility guidelines and auditable checkpoints. 

\subsection{Security and Privacy}

Due to the decentralized nature of AAI, security and privacy risks can be observed across edge devices, cloud systems, and communication networks. Agents are able to process data in real time from a variety of data sources, making them a potential target for attacks, data breaches and misinformation campaigns. Malicious actors can attempt to sabotage the AI agents by exploiting vulnerabilities. For example, a malicious attacker may feed in false data, leading to incorrect disaster assessment or suboptimal resource allocation. In addition, security and privacy concerns arise when AI agents share sensitive data such as geolocation and personal identifiers with emergency responders and other AI agents. Ensuring security and privacy through blockchain-based trust mechanisms and encryption would be critical in the AAI-based frameworks.

\section{Conclusion}

In this paper, we present an AAI framework for mission-critical public safety applications. We provide a layered architecture for the AAI-based mission-critical applications and also provide a detailed AAI layer block that connects both the network infrastructure layer and the mission-critical application layer. We also present an experimental setup and analysis to show the realization of a part of the AAI framework and prove its efficiency and effectiveness. We show that the AAI approach performs significantly better than the non-AAI approaches on a number of features and performance metrics. Although the AAI approach is effective, there are challenges and issues that need to be addressed before implementing the AAI framework. We potentially address three of these challenges, namely the openness of foundational models, trust \& accountability, and security \&privacy. \\
In the future, the network bandwidth and memory utilisation of the AAI framework can be improved  by adding rewards when training the agents. In addition,  the agents that are sequentially trained  can  be extended to perform multi-agent collaboration. Therefore, adding more agents and training them in a collaborative way while utilising the Retrieval Augmented Generation (RAG) functionalities may be another future direction to pursue. 

\balance

% \newpage

 

% \end{IEEEbiography}


% \bibliographystyle{ieee} % or any other style you prefer
% \bibliography{biblio} % without the .bib extension

\begin{thebibliography}{1}
\bibliographystyle{IEEEtran.bst}

\bibitem{6GMC}
A. F. M. Shahen Shah, M. Ali Karabulut and K. Rabie, "Mission-Critical Internet of Things on the 6G Network: Services and Apps with Networking Architecture," 2023 IEEE 98th Vehicular Technology Conference (VTC2023-Fall), Hong Kong, Hong Kong, 2023, pp. 1-5, doi: 10.1109/VTC2023-Fall60731.2023.10333821.

\bibitem{SmartMC}
S. T. Spantideas, A. E. Giannopoulos and P. Trakadas, "Smart Mission Critical Service Management: Architecture, Deployment Options, and Experimental Results," in IEEE Transactions on Network and Service Management, doi: 10.1109/TNSM.2024.3498348.

\bibitem{IMC5G}
S. Spantideas, A. Giannopoulos, M. A. Cambeiro, O. Trullols-Cruces, E. Atxutegi and P. Trakadas, "Intelligent Mission Critical Services over Beyond 5G Networks: Control Loop and Proactive Overload Detection," 2023 International Conference on Smart Applications, Communications and Networking (SmartNets), Istanbul, Turkiye, 2023, pp. 1-6, doi: 10.1109/SmartNets58706.2023.10216134.

\bibitem{AIMC}
Perez-Cerrolaza, J., Abella, J., Borg, M., Donzella, C., Cerquides, J., Cazorla, F.J., Englund, C., Tauber, M., Nikolakopoulos, G. and Flores, J.L., 2024. Artificial intelligence for safety-critical systems in industrial and transportation domains: A survey. ACM Computing Surveys, 56(7), pp.1-40.

\bibitem{AgenticAI}
Dev, K., Khowaja, S.A., Zeydan, E. and Debbah, M., 2025. Advanced Architectures Integrated with Agentic AI for Next-Generation Wireless Networks. arXiv preprint arXiv:2502.01089.

\bibitem{6GAI}
Chataut, R., Nankya, M. and Akl, R., 2024. 6G networks and the AI revolution—Exploring technologies, applications, and emerging challenges. Sensors, 24(6), p.1888.

\bibitem{Wireless6G}
A. E. Kalor et al., "Wireless 6G Connectivity for Massive Number of Devices and Critical Services," in Proceedings of the IEEE, doi: 10.1109/JPROC.2024.3484529.

\bibitem{6Gedge}
Gupta, R., Reebadiya, D. and Tanwar, S., 2021. 6G-enabled edge intelligence for ultra-reliable low latency applications: Vision and mission. Computer Standards \& Interfaces, 77, p.103521.

\bibitem{XAI6G}
S. Wang, M. A. Qureshi, L. Miralles-Pechuán, T. Huynh-The, T. R. Gadekallu and M. Liyanage, "Explainable AI for 6G Use Cases: Technical Aspects and Research Challenges," in IEEE Open Journal of the Communications Society, vol. 5, pp. 2490-2540, 2024, doi: 10.1109/OJCOMS.2024.3386872

\bibitem{DBFL}
S. A. Khowaja, K. Dev, P. Khowaja and P. Bellavista, "Toward Energy-Efficient Distributed Federated Learning for 6G Networks," in IEEE Wireless Communications, vol. 28, no. 6, pp. 34-40, December 2021, doi: 10.1109/MWC.012.2100153.

\bibitem{PublicSafetyLTE}
K. Muraoka, J. Shikida and H. Sugahara, "Feasibility of capacity enhancement of public safety LTE using device-to-device communication," 2015 International Conference on Information and Communication Technology Convergence (ICTC), Jeju, Korea (South), 2015, pp. 350-355, doi: 10.1109/ICTC.2015.7354561.

\bibitem{FirstNet}
C. Budny, J. Liu and A. Weinert, "Video Testing at the FirstNet Innovation and Test Lab Using a Public Safety Dataset," in IEEE Networking Letters, vol. 2, no. 1, pp. 28-32, March 2020, doi: 10.1109/LNET.2019.2955833.

\bibitem{Transformer}
Waswani, A., Shazeer, N., Parmar, N., Uszkoreit, J., Jones, L., Gomez, A., Kaiser, L. and Polosukhin, I., 2017, December. Attention is all you need. In NIPS.


\bibitem{Crisis}
Olteanu, A., Castillo, C., Diaz, F. and Vieweg, S., 2014, May. Crisislex: A lexicon for collecting and filtering microblogged communications in crises. In Proceedings of the international AAAI conference on web and social media (Vol. 8, No. 1, pp. 376-385).

\bibitem{Ulema}
Ulema, Mehmet. Fundamentals of Public Safety Networks and Critical Communications Systems: Technologies, Deployment, and Management. John Wiley \& Sons, 2019.

\end{thebibliography}

\vfill

\end{document}